# Hierarchical Spatial Proximity Reasoning for Vision-and-Language Navigation

Ming Xu, Zilong Xie

*Abstract*—Most Vision-and-Language Navigation (VLN) algorithms are prone to making inaccurate decisions due to their lack of visual common sense and limited reasoning capabilities. To address this issue, we propose a Hierarchical Spatial Proximity Reasoning (HSPR) method. First, we introduce a scene understanding auxiliary task to help the agent build a knowledge base of hierarchical spatial proximity. This task utilizes panoramic views and object features to identify types of nodes and uncover the adjacency relationships between nodes, objects, and between nodes and objects. Second, we propose a multi-step reasoning navigation algorithm based on the hierarchical spatial proximity knowledge base, which continuously plans feasible paths to enhance exploration efficiency. Third, we introduce a residual fusion method to improve navigation decision accuracy. Finally, we validate our approach with experiments on publicly available datasets including REVERIE, SOON, R2R, and R4R. Our code is available at *https://github.com/iCityLab/HSPR*

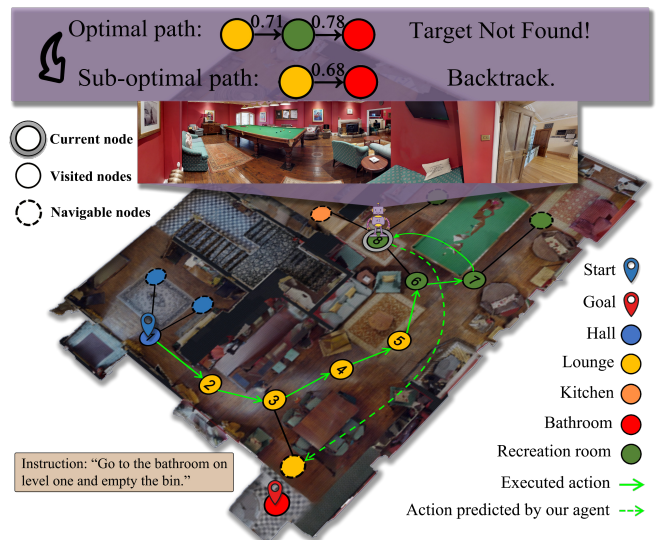

**Fig. 1.** An example of reasoning navigation. In the semantic topological map, nodes are categorized as the current node, visited nodes, and navigable nodes, with different colors representing each type. Based on the proximity probabilities between the various node types and the target, the agent infers two feasible paths. It first explores the optimal path but discovers that there is no bathroom near the recreation room. Consequently, it switches to the sub-optimal path, backtracks, and eventually reaches the target.

## I. INTRODUCTION

IN recent years, Vision-and-Language Navigation (VLN) [1] has played a crucial role in various domains of embodied AI tasks, including social services, rescue operations, and human-computer interaction. VLN aims to enable agents to understand human instructions, recognize visual observations in the real world, and achieve precise navigation in unseen environments. Based on the characteristics of language instructions, VLN can be categorized into fine-grained navigation that provides step-by-step instructions (e.g., R2R [1]), goal-oriented coarse-grained navigation (e.g., REVERIE [2]), and dialogue-based navigation with human-agent interaction (e.g., DialFRED [3]). Among these, coarse-grained navigation, with its concise instruction characteristics, can adapt to a wider range of application scenarios.

Early research [1, 2, 4, 5, 6] focused on fine-grained navigation, employing recurrent neural network (such as LSTM [7]) to construct agents. However, due to inefficiencies in learning long-term dependencies, these models struggled to propagate navigation history information during recurrent steps. Recent studies [8–11] have addressed coarse-grained navigation using transformer [12] architecture. These approaches leverage self-attention mechanisms to model long-term dependencies between navigation actions and introduce topological maps for global and local planning. However, they do not explicitly incorporate spatial proximity knowledge, which can lead to agents exploring blindly during navigation.

Some studies [13, 14] have attempted to address this issue by utilizing concept-based or image-based knowledge bases [15, 16]. However, these approaches have two drawbacks: 1) The knowledge bases are not specifically tailored for navigation tasks leading to an excess of irrelevant knowledge that reduces the efficiency of knowledge utilization. 2) These methods do not fully leverage proximity knowledge to enhance decision-making capabilities. For example, they consider only proximity between the current node and the target, neglecting proximity to intermediate transit nodes.

To address the aforementioned issues, we propose a Hierarchical Spatial Proximity Reasoning (HSPR) method, as

Ming Xu and Zilong Xie are with Software College, Liaoning Technical University, Huludao, China (e-mail: xum.2016@tsinghua.org.cn, zilong6037@gmail.com).

illustrated in Fig. 1. This method includes two stages: pretraining and navigation. In the pretraining stage, we introduce an auxiliary task to construct a hierarchical spatial proximity knowledge base and train the agent to accurately identify node types within the navigation environment. During the navigation stage, we utilize a multi-step reasoning navigation algorithm based on this proximity knowledge base to plan the top-*K* feasible paths, exploring them in a prioritized manner to efficiently reach the target. Additionally, we propose a residual fusion method that fuses visual information from navigable nodes with their spatial proximity knowledge, enabling more accurate navigation decisions.

**Contributions:** 1) We introduce a scene understanding auxiliary task that enables the agent to construct a hierarchical spatial proximity knowledge base from the navigation environment. 2) We propose a multi-step reasoning navigation algorithm based on the hierarchical spatial proximity knowledge base, achieving efficient exploration and navigation decision-making. 3) We validate the effectiveness of the proposed method through experiments on the VLN benchmark datasets. In the coarse-grained navigation datasets REVERIE and SOON, our method improved the navigation success rate (SR) by at least 2.07% and 5.90% over baseline methods, respectively. Additionally, the success weighted by path length (SPL) increased by at least 0.57% and 5.90%, respectively.

## II. RELATED WORK

**Visual and Language Navigation.** Progress in VLN has been achieved through innovations at various levels, including from fine-grained [1, 17, 18] to coarse-grained [2, 19] instruction conversion, memory structure improvements from state vectors [5, 6] to topological maps [9, 20, 21], architectural advancements from recurrent neural networks with cross-modal attention [1, 2, 4–6] to transformers [8–11, 13, 14], and continuous innovation in representation learning [8, 22, 23], navigation strategies [24, 25, 26], exploration and exploitation [27], and data augmentation [6, 28, 29]. Recently, most VLN methods [8, 23, 30, 31] have relied on large-scale pretrained networks to encode visual and language features. However, these methods did not explicitly utilize spatial proximity cues beyond the two input modalities, which are crucial for guiding agents.

**Auxiliary Tasks for VLN.** Auxiliary tasks have played a crucial role in assisting agents to better understand the environment and their own state, leading to more efficient and robust predictions for future decision-making [5, 32, 33]. For example, AuxRN [5] enhances the internal state representation of the policy network through auxiliary tasks, improving the agent's performance. SEA [32] improves the visual representation of the environment by refining the image encoder through an auxiliary task. Auxiliary tasks also include monitoring task completion [34] and alignment between visual and navigation instructions [29]. Our proposed auxiliary task effectively uncovers adjacency relationships between nodes, objects, and between nodes and objects in the environment. These adjacency relationships are then applied in subsequent navigation tasks, leading to a significant improvement in navigation efficiency.

**Vision-Language Reasoning.** The vision-language multi-modal tasks, which connect computer vision and natural language processing, have attracted significant attention, with visual-language reasoning playing a crucial role in these tasks. Recent studies [30, 35] have focused on obtaining higher-level semantics by pretraining models on large-scale datasets with visual language reasoning tasks. Other studies [4, 36] have employed cross-modal attention mechanisms to reason about visual entities. Additionally, some studies combine external knowledge bases for reasoning in navigation. For example, CKR [13] retrieves graph-structured knowledge related to room and object entities from ConceptNet [15], while KERM [14] leverages a knowledge base built from the Visual Genome dataset [16] to retrieve facts about navigation views, enabling a better understanding of visual concepts. In contrast, we extract knowledge directly from the navigation environment and use it for reasoning in the navigation task, without relying on external knowledge bases.

## III. METHOD

In this section, we introduce the Hierarchical Spatial Proximity Reasoning (HSPR) method. In the VLN task within discrete environment, a navigation scene is represented as an undirected graph $\mathcal{G} = \{\mathcal{N}, E\}$, where $\mathcal{N}$ denotes the set of $|\mathcal{N}|$ nodes and $E$ represents the connecting edges. Each node $n$ is associated with a panoramic view $\mathcal{P} = \{p_i\}_{i=1}^{36}$, which is divided into 36 parts. The type of each node refers to the type of region in which the node is located, as determined through scene semantic segmentation. Each part $p_i = \{v_i, o_i, \phi_i, \varphi_i\}$ contains an image representation $v_i$, an object representation $o_i$, a heading angle $\phi_i$, and an elevation angle $\varphi_i$. Navigable nodes are those that have been locally observed from previously visited nodes but have not yet been physically traversed by the agent. The agent starts at a random node and receives an instruction $I = \{w_j\}_{j=1}^{J}$, where $J$ is the number of words in the instruction. At each step $t$, the agent observes a panoramic view of the current node. Additionally, the agent is aware of its navigable nodes and their corresponding view set. The agent must then take an action, selecting a node from the set of navigable nodes $\{n_{t,i}\}_{i=0}^{K_t}$, where $K_t$ is the number of navigable nodes and $n_{t,0}$ represents the stop action. Once the agent decides to stop at a particular node, it needs to predict the location of the target object in the panoramic view.

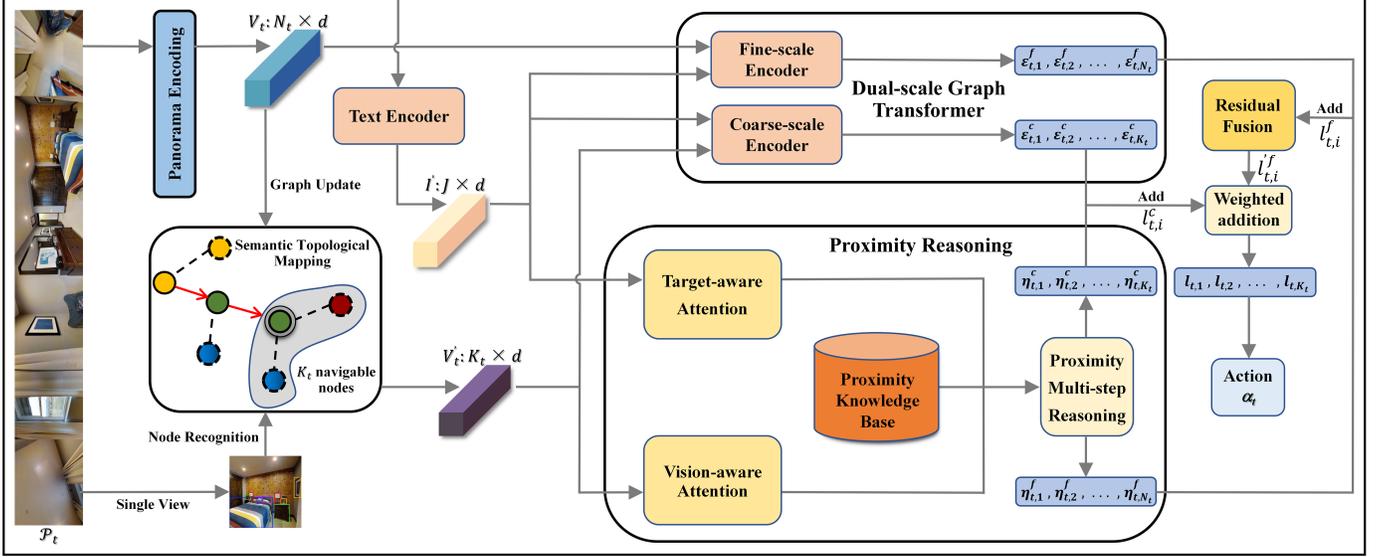

**Fig. 2.** The overview of the proposed HSPR.

Our proposed model is presented in Fig. 2. The text encoder processes the navigation instructions. The semantic topological mapping module incrementally constructs and updates a semantic topological map. The proximity reasoning module leverages proximity knowledge acquired from the scene understanding auxiliary task, employing a multi-step reasoning navigation algorithm to determine sub-goals and calculate the node proximity score for each navigable node corresponding to the sub-goals. The residual fusion then integrates these scores with the visual scores predicted by the Dual-scale Graph Transformer (DUET) [11], resulting in more accurate navigation action.

*A. Scene Understanding Auxiliary Task*

To obtain common sense information and accurately predict node types for navigation, we propose a scene understanding auxiliary task. This task records the adjacency relationships in the navigation environment and trains the agent to recognize the node types.

**Acquiring Proximity Knowledge.** We first utilize the node types and object labels in the navigation scenes to obtain the node adjacency count matrix $C^r \in \mathbb{R}^{N_r \times N_r}$, the object adjacency count matrix $C^o \in \mathbb{R}^{N_o \times N_o}$, and the node-object correlation count matrix $C^{ro} \in \mathbb{R}^{N_r \times N_o}$, where $N_r$ represents the number of node types and $N_o$ represents the number of object types. Specifically, we segment each scene into different regions. If there are multiple disconnected regions of the same type, we further divide them into independent regions. We then count the connectivity between these segmented regions, which is considered the connections between different types of nodes. In each view $p_i$, all objects are regarded as adjacent, resulting in an increment of their co-occurrence count by one.

Next, to obtain the proximity probability matrix, we replace extreme high values in each row of the matrix with the 95th percentile, followed by Min-Max normalization to convert the processed proximity count matrix into a proximity probability matrix $P$ with a range of [0, 0.95]:

$$P_{i,j} = \frac{0.95 \times (C_{i,j} - C_i^{min})}{C_i^{max} - C_i^{min}} \quad (1)$$

where $C_i^{min}$ is the minimum value of the $i$-th row of matrix $C$ after outlier treatment, and $C_i^{max}$ corresponds to the maximum value. Finally, to extract spatial co-occurrence knowledge between nodes and objects, we identify the $K$ objects that co-occur most frequently with each node.

**Node Recognition.** This task involves predicting the type of a node using its corresponding panoramic view. At each step $t$, the agent receives the visual features $v_{t,i}$ of the $i$-th navigable node and the features $o_{t,i}$ of the objects it contains. Following the method in [11], we use the self-attention (SA) block [12] to model the spatial relationships between visual and object features, enhancing their representations. The enhanced visual features $v'_{t,i}$ are then fed into a fully connected (FC) layer to compute the type scores $s_{t,i} \in \mathbb{R}^{1 \times N_r}$ for the $i$-th navigable node:

$$\text{SA}(x;\theta) = \sigma\left(d^{-\frac{1}{2}} x W_Q (x W_K)^\mathrm{T}\right) x W_V \quad (2)$$

$$[v'_{t,i} | o'_{t,i}] = \text{SA}([v_{t,i} | o_{t,i}]; \theta) \quad (3)$$

$$s_{t,i} = \sigma(\text{FC}(v'_{t,i})) \quad (4)$$

where $\sigma$ is the softmax activation function, the operator '|' represents vector concatenation, $W_Q$, $W_K$ and $W_V \in \mathbb{R}^{d \times d}$ are learnable parameters, collectively denoted as $\theta$. We still use $v_{t,i}, o_{t,i}$ instead of $v'_{t,i}, o'_{t,i}$ to represent the encoded embedding in the following sections. The loss function used for node classification is cross-entropy:

$$L_R = \sum_{t=1}^{T} \sum_{i=1}^{N_t} (-\log s_{t,i}) \quad (5)$$

where $T$ is the total number of steps in the navigation task.

*B. Proximity Reasoning*

To better utilize the hierarchical spatial proximity knowledge, we design the proximity reasoning module. Specifically, we use the attention mechanism to predict the types of navigable nodes and the target node from the instructions. We then employ a multi-step reasoning algorithm to carry out the navigation process.

**Proximity Calculating.** The instruction embedding $I'$, obtained through the text encoder, is initially input into a target-aware attention layer to predict the type $Y^r$ of the target node and the type $Y^o$ of target object, which is formulated as:

$$Y^r, Y^o = \sigma\left(\text{FC}\left(\text{SA}(I'; \theta_{ta})\right)\right) \quad (6)$$

At each step $t$, the visual representation $v_{t,i}$ of $i$-th navigable node is fed into the vision-aware attention layer to determine the node type $R_{t,i}$ via:

$$R_{t,i} = \sigma\left(\text{FC}\left(\text{SA}(v_{t,i}; \theta_{va})\right)\right) \quad (7)$$

Once the stop action is selected, the object representations $o_{T,0}$ of the stop node is fed into the vision-aware attention layer, to predict the object types $O_{T,0}$. The formula is as:

$$O_{T,0} = \sigma\left(\text{FC}\left(\text{SA}(o_{T,0}; \theta'_{va})\right)\right) \quad (8)$$

Based on these, we leverage the hierarchical spatial proximity knowledge to compute the node proximity scores $\eta_{t,i}$ between the navigable nodes and the target at each step, as well as the object proximity scores $\mu_{T,0}$ between various objects and the target object at step $T$:

$$\eta_{t,i} = R_{t,i} P^r Y^{rT}, \quad \mu_{T,0} = O_{T,0} P^o Y^{oT} \quad (9)$$

where $P^r$ and $P^o$ represent the node and object proximity probability matrices, respectively, obtained from the scene understanding auxiliary task.

**Multi-step Reasoning Navigation Algorithm.** The initial position of the agent is often not adjacent to the target, resulting in low or even zero proximity scores for some navigable nodes. However, the agent can utilize these nodes as intermediate stations to gradually approach the target until it successfully reaches the destination.

Building on this intuition, we propose a multi-step reasoning navigation algorithm that allows the agent to continuously plan multiple feasible paths from one node to another, as illustrated in Fig. 3. At each step $t$, the agent considers the top-$K$ candidate paths with the highest confidence and selects the optimal path. As the semantic topological map is gradually constructed, new types of nodes are continuously added, and the agent must ensure that the current path remains optimal. In the estimated paths, the next intermediate node is treated as a sub-goal. If no sub-goal is found, the agent selects the sub-optimal path until the target is reached. Considering the intermediate nodes, the proximity

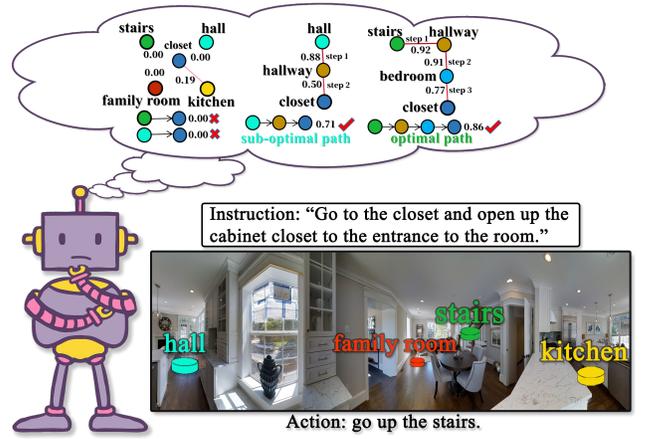

Fig. 3. Multi-step reasoning navigation example. Despite the low probabilities of the current navigable nodes leading directly to the target node, the intelligent agent utilizes the multi-step reasoning navigation algorithm to plan multiple feasible paths. From these paths, the agent selects the optimal path, which involves going through the staircase, hallway, and bedroom to reach the target node.

score $\eta'_{t,i}$ of the $i$-th navigable node to the target is re-calculated as follows:

$$\eta'_{t,i} = \gamma^0 \omega_1 \eta^{(1)}_{t,i} + \gamma^1 \omega_2 \eta^{(2)}_{t,i} + \cdots + \gamma^{m-1} \omega_m \eta^{(m)}_{t,i} \quad (10)$$

where $\gamma$ is the discount factor, $\omega_i$ ($i = 1, 2, \ldots, m$) is the weight coefficient, $\eta^{(m)}_{t,i}$ represents the node score towards the sub-goal obtained from the $m$-th step proximity calculation.

*C. Residual Fusion*

In this work, we consider all navigable nodes as the global navigable node set $\mathcal{C}$, while the local navigable node set $\mathcal{F}$ includes only the navigable nodes adjacent to the current node. The complement of $\mathcal{F}$ within $\mathcal{C}$ is referred to as the non-local navigable nodes. We use DUET to predict the global and local visual scores ($\varepsilon^c_{t,i}$ and $\varepsilon^f_{t,i}$), which are combined with the proximity scores ($\eta^c_{t,i}$ and $\eta^f_{t,i}$), to produce the global and local action scores ($l^c_{t,i}$ and $l^f_{t,i}$). Our residual fusion method converts local action scores $l^f_{t,i}$ into global action scores $l^c_{t,i}$. Unlike dynamic fusion [11], which inaccurately sums the action scores of all the visited nodes for the non-local navigable nodes. In contrast, residual fusion combines the global node proximity score and visual score for each global navigable node and assigns them to the non-local navigable node:

$$l'^f_{t,i} = \begin{cases} \eta^f_{t,i}, & \text{if } n_{t,i} \in \mathcal{F}, \\ \eta^c_{t,i} + \varepsilon^c_{t,i}, & \text{otherwise}. \end{cases} \quad (11)$$

We utilize the local and global stop action embeddings $v^f_{t,0}$ and $v^c_{t,0}$ encoded by the local-global cross-modal attention modules [11] to predict a balancing factor $\beta_t$ as:

$$\beta_t = \text{Sigmoid}\left(\text{FFN}([v^c_{t,0} | v^f_{t,0}])\right) \quad (12)$$

TABLE I
COMPARISON WITH SOTA METHODS ON REVERIE DATASET

| Methods | Val Seen | | | | | | Val Unseen | | | | | | Test Unseen | | | | | |
|---|---|---|---|---|---|---|---|---|---|---|---|---|---|---|---|---|---|---|
| | Navigation | | | | Grounding | | Navigation | | | | Grounding | | Navigation | | | | Grounding | |
| | TL | OSR↑ | SR↑ | SPL↑ | RGS↑ | RGSPL↑ | TL | OSR↑ | SR↑ | SPL↑ | RGS↑ | RGSPL↑ | TL | OSR↑ | SR↑ | SPL↑ | RGS↑ | RGSPL↑ |
| Human | - | - | - | - | - | - | - | - | - | - | - | - | 21.18 | 86.83 | 81.51 | 53.66 | 77.84 | 51.44 |
| Seq2Seq [1] | 12.88 | 35.70 | 29.59 | 24.01 | 18.97 | 14.96 | 11.07 | 8.07 | 4.20 | 2.84 | 2.16 | 1.63 | 10.89 | 6.88 | 3.99 | 3.09 | 2.00 | 1.58 |
| SMNA [34] | 7.54 | 43.29 | 41.25 | 39.61 | 30.07 | 28.98 | 9.07 | 11.28 | 8.15 | 6.44 | 4.54 | 3.61 | 9.23 | 8.39 | 5.80 | 4.53 | 3.10 | 2.39 |
| CKR [13] | 12.16 | 61.91 | 57.27 | 53.57 | 39.07 | - | 26.26 | 31.44 | 19.14 | 11.84 | 11.4 | - | 22.46 | 30.40 | 22.00 | 14.25 | 11.60 | - |
| VLNBERT [23] | 13.44 | 53.90 | 51.79 | 47.96 | 38.23 | 35.61 | 16.78 | 35.02 | 30.67 | 24.90 | 18.77 | 15.27 | 15.68 | 32.91 | 29.61 | 23.99 | 16.50 | 13.51 |
| HAMT [10] | 12.79 | 47.65 | 43.29 | 40.19 | 27.20 | 15.18 | 14.08 | 36.84 | 32.95 | 30.20 | 18.92 | 17.28 | 13.62 | 33.41 | 30.40 | 26.67 | 14.88 | 13.08 |
| Airbert [31] | 15.16 | 49.98 | 47.01 | 42.34 | 32.75 | 30.01 | 18.71 | 34.51 | 27.89 | 21.88 | 18.23 | 14.18 | 17.91 | 34.20 | 30.28 | 23.61 | 16.83 | 13.28 |
| DUET [11] | 13.86 | 73.68 | 71.75 | 63.94 | 57.41 | 51.14 | 22.11 | 51.07 | 46.98 | 33.73 | 32.15 | 23.03 | 21.30 | 56.91 | 52.51 | 36.06 | 31.88 | 22.06 |
| KERM [14] | 12.84 | 79.20 | 76.88 | **70.45** | 61.00 | 56.07 | 21.85 | 55.21 | 50.44 | 35.38 | 34.51 | 24.45 | 17.32 | 57.58 | 52.43 | 39.21 | 32.39 | 23.64 |
| STPR (Ours) | 13.69 | **80.84** | **77.51** | 70.32 | **62.49** | **56.53** | 22.97 | **58.85** | **52.94** | **38.05** | **36.84** | **26.51** | 19.06 | **60.24** | **54.50** | **39.78** | **34.60** | **24.07** |

TABLE II
COMPARISON WITH SOTA METHODS ON SOON DATASET.
† INDICATES REPRODUCED RESULTS

| Methods | Val Seen House | | | | Unseen House (Test) | | | |
|---|---|---|---|---|---|---|---|---|
| | OSR↑ | SR↑ | SPL↑ | RGSPL↑ | OSR↑ | SR↑ | SPL↑ | RGSPL↑ |
| Human | - | - | - | - | 91.4 | 90.4 | 59.2 | 51.1 |
| AuxRN [5] | 78.5 | 68.8 | 67.3 | 8.3 | 11.0 | 8.1 | 6.7 | 0.5 |
| GBE w/o GE | 73.0 | 62.5 | 60.8 | 6.7 | 18.8 | 11.4 | 8.7 | 0.8 |
| GBE [19] | 64.1 | 76.3 | 62.5 | 7.3 | 19.5 | 11.9 | 10.2 | 1.4 |
| DUET† | 80.3 | 78.2 | 69.5 | 11.9 | 43.0 | 33.4 | 21.4 | 4.2 |
| STPR(Ours) | **84.8** | **81.0** | **70.6** | **12.5** | **53.7** | **39.3** | **27.3** | **4.7** |

TABLE III
COMPARISON WITH SOTA METHODS ON R2R DATASET

| Methods | Val Unseen | | | | Test Unseen | | | |
|---|---|---|---|---|---|---|---|---|
| | TL | NE↓ | SR↑ | SPL↑ | TL | NE↓ | SR↑ | SPL↑ |
| Seq2Seq [1] | 8.93 | 7.81 | 21 | - | 8.13 | 7.85 | 20 | 18 |
| AuxRN [5] | - | 5.28 | 54 | 50 | - | 5.15 | 55 | 51 |
| SSM [21] | 20.7 | 4.32 | 62 | 45 | 20.4 | 4.57 | 61 | 46 |
| VLNBERT [23] | 12.01 | 3.93 | 63 | 57 | 12.35 | 4.09 | 63 | 57 |
| HAMT-e2e [10] | 11.46 | **2.29** | 66 | 61 | 12.27 | 3.93 | 65 | 60 |
| DUET [11] | 13.94 | 3.31 | 71.52 | 60.42 | 14.73 | 3.65 | 69.25 | 58.68 |
| KERM [14] | 13.54 | 3.22 | 71.95 | 60.91 | 14.60 | 3.61 | 69.73 | 59.25 |
| STPR(Ours) | 13.83 | 3.10 | **73.27** | **62.45** | 15.67 | **3.38** | **70.57** | **59.34** |

TABLE IV
COMPARISON WITH SOTA METHODS ON R4R DATASET. †
INDICATES REPRODUCED RESULTS

| Methods | Val Seen | | | | Val Unseen | | | |
|---|---|---|---|---|---|---|---|---|
| | TL | NE↓ | SR↑ | SPL↑ | TL | NE↓ | SR↑ | SPL↑ |
| Speaker-Follower [6] | 15.40 | 5.35 | 52 | 37 | 19.90 | 8.47 | 24 | 12 |
| EnvDrop [28] | 19.85 | - | 52 | 41 | 26.97 | - | 29 | 18 |
| OAAM [40] | 11.75 | - | 56 | 49 | 13.80 | - | 31 | 23 |
| SSM [21] | 19.40 | 4.60 | 63 | - | 22.10 | **3.22** | 32 | - |
| DUET† | 18.63 | 4.38 | 63.00 | 60.11 | 23.02 | 6.40 | 45.11 | 40.62 |
| STPR(Ours) | 17.06 | **4.17** | **64.38** | **61.67** | 19.63 | 6.02 | **46.94** | **43.96** |

where FFN denotes a two-layer feed-forward network.

Finally, we combine the local action scores from the residual fusion with the global action scores by applying a weighted sum to obtain the final action score $l_{t,i}$:

$$l_{t,i} = \beta_t l_{t,i}^c + (1-\beta_t) l_{t,i}'^f \quad (13)$$

### D. Training and Inference

**Pre-training.** In previous studies [8, 23, 30, 31, 35], it has been demonstrated that pretraining techniques significantly improve model performance. To pretrain the proposed VLN model, this study adopts Masked Language Modeling (MLM) [37] for language encoder pretraining. Additionally, Masked Region Classification (MRC) [30], Single-step Action Prediction (SAP) [10], and Object Grounding (OG) [38] are used to enhance the visual encoder. The SAP and OG losses given a demonstration path $Q^*$ are defined as follows:

$$L_{SAP} = \sum_{t=1}^{T} -\log p(a_t^*|I, Q_{<t}^*) \quad (14)$$

$$L_{OG} = -\log p(o^*|I, n_T) \quad (15)$$

where $a_t^*$ represents the ground truth action for the partial demonstration path $Q_{<t}^*$, and $o^*$ is the ground truth of object within the last node $n_T$.

**Fine-tuning and Inference.** We introduce the Pseudo Interactive Demonstrator (PID) [11] to further train the policy, aiming to mitigate the impact of distribution shifts between training and testing. The loss function $L_{PID}$ follows the same structure as the SAP loss in pre-training. However, $a_t^*$ is replaced with the pseudo-truth action, which selects the navigable node with the shortest total distance from the current node to the destination. Additionally, the demonstrated path $Q^*$ is replaced with a trajectory sampled based on the current learned policy. The total loss function is defined as $L_{TOTAL} = 0.2L_{SAP} + L_{PID} + L_{OG} + L_R$.

During the navigation process, the model determines the action to be taken at each step. The agent uses the Floyd algorithm, which is suitable for medium-sized maps, to plan the optimal path on the constructed semantic topological map and follows this path to the navigable node. This process continues until a stop action is selected or the maximum number of steps is exceeded. Once the agent stops at a certain node, it selects the highest-scoring object as the target.

## IV. EXPERIMENTS

### A. Datasets

We evaluate our model on four VLN benchmark datasets: R2R [1], R4R [17], REVERIE [2], and SOON [19], all of which are split into training, validation and testing sets. The validation set is further divided into a validation-seen set (val-seen) and a validation-unseen set (val-unseen). The former includes new paths and instructions, but the navigation scenes are the same as those seen in the training set, while the latter contains entirely new paths, instructions, and navigation scenes. REVERIE gives high-level instructions, requiring the agent to select the bounding box of the target object after stopping. SOON provides location information for objects near the target and the room in which the target is located. R2R offers detailed step-by-step instructions. while R4R includes longer paths.

TABLE V
ABLATIONS OF PROXIMITY KNOWLEDGE AND
UTILIZATION ON REVERIE VAL UNSEEN SPLIT

| Methods | OSR↑ | SR↑ | SPL↑ | RGS↑ | RGSPL↑ |
|---|---|---|---|---|---|
| BaseNet | 51.07 | 46.98 | 33.73 | 32.15 | 23.03 |
| +ConceptNet | 55.13 | 50.10 | 35.59 | 34.00 | 24.58 |
| +Auxiliary task | **58.85** | **52.94** | **38.05** | **36.84** | **26.51** |

TABLE VI
THE EFFECT OF REASONING STEP NUMBER IN MULTI-STEP REASONING NAVIGATION ALGORITHM ON REVERIE VAL UNSEEN SPLIT

| Number of steps | OSR↑ | SR↑ | SPL↑ | RGS↑ | RGSPL↑ |
|---|---|---|---|---|---|
| 1 | 56.72 | 51.66 | 36.13 | 34.73 | 24.28 |
| 2 | **59.02** | **53.14** | 37.07 | 36.68 | 25.75 |
| 3 | 58.85 | 52.94 | **38.05** | **36.84** | **26.51** |
| 4 | 57.71 | 52.14 | 37.79 | 35.36 | 25.85 |
| 5 | 56.21 | 50.78 | 35.46 | 34.56 | 24.14 |

TABLE VII
COMPARE THE PERFORMANCE OF THREE DIFFERENT FUSION METHODS ON REVERIE VAL UNSEEN SPLIT

| Fusion | OSR↑ | SR↑ | SPL↑ | RGS↑ | RGSPL↑ |
|---|---|---|---|---|---|
| Average | 57.14 | 50.87 | 35.08 | 33.80 | 23.13 |
| Dynamic | 56.57 | 52.07 | 37.21 | 35.59 | 25.62 |
| Residual | **58.85** | **52.94** | **38.05** | **36.84** | **26.51** |

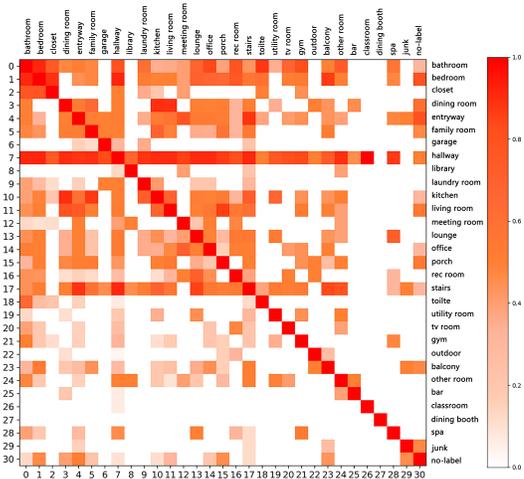

**Fig. 4.** Proximity knowledge of nodes constructed from auxiliary tasks.

*B. Evaluation Metrics*

We use standard metrics [1] to measure navigation performance, including Trajectory Length (TL): the average path length in meters, Navigation Error (NE): the average distance between the final position of agent and the target in meters, Success Rate (SR): the proportion of paths with NE less than 3 meters, Oracle Success Rate (OSR): SR given an oracle stop policy, Success weighted by Path Length (SPL): SR penalized by path length. To evaluate object grounding, we use Remote Grounding Success (RGS) [2]: the proportion of successfully executed instructions, RGS weighted by Path Length (RGSPL): RGS penalized by path length. Except for TL and NE, higher values are better for all metrics.

*C. Model Architectures*

The dimensions of the node proximity count matrix $N_r$ and the object proximity count matrix $N_o$ are 31 and 1600, respectively. The dimensions of the feature channels $d$ is set to 768. The number of transformer layers for instructions, panorama, and local-global cross-modal attention modules are set to 9, 2, 4 and 4, respectively. The discount factor $\gamma$ for multi-step reasoning is set to 0.9. We initialize the model using the pre-trained LXMERT [36] and utilize the pre-trained ViT-B/16 [39] to extract image and object features. For the SOON dataset, we employ the BUTD object detector [4] for object detection.

*D. Comparison with SOTA Methods*

Table I shows the results of HSPR on the REVERIE dataset. Compared with the best baseline, HSPR improves SR and SPL by 2.50% and 2.67%, respectively, on the val-unseen split, and by 2.07% and 0.57% on the test unseen split. As shown in Table II, for SOON dataset, HSPR significantly outperforms other baselines on the test unseen split, with SR increasing by 5.90% and SPL by 5.90%. These results suggest that HSPR can effectively enhance reasoning-based navigation performance by leveraging proximity knowledge without detailed step-by-step instruction guidance.

Table III presents the results on the R2R dataset. Compared with the best baselines, HSPR improves SR by 1.32% and SPL by 1.54% on the val-unseen split and increases SR by 0.84% and SPL by 0.09% on the test-unseen split. As shown in Table IV, on the R4R dataset, HSPR shows improvements in SR by 1.38% and 1.83%, and in SPL by 1.56% and 3.34% on the val-seen and val-unseen splits, respectively. HSPR's performance on R2R is significantly lower than that on REVERIE, though it continues to perform well on R4R. We attribute this to the longer navigation paths in R4R, which enable the multi-step reasoning navigation algorithm to demonstrate its capabilities.

*E. Ablation Study*

We conducted ablation experiments on the REVERIE dataset to investigate the contributions of each component to navigation performance. All the results in this section are reported on the val-unseen split.

**Proximity knowledge.** We compared the impact of proximity knowledge obtained from different methods on navigation performance. As shown in Table V, using proximity knowledge retrieved from the ConceptNet [15] outperforms the baseline, which lacks any proximity knowledge. Our proposed method, which leverages the scene understanding auxiliary task to obtain proximity knowledge, performs the best. This indicates that proximity knowledge can indeed improve navigation performance, and that knowledge obtained from our auxiliary task is more suitable for navigation tasks.

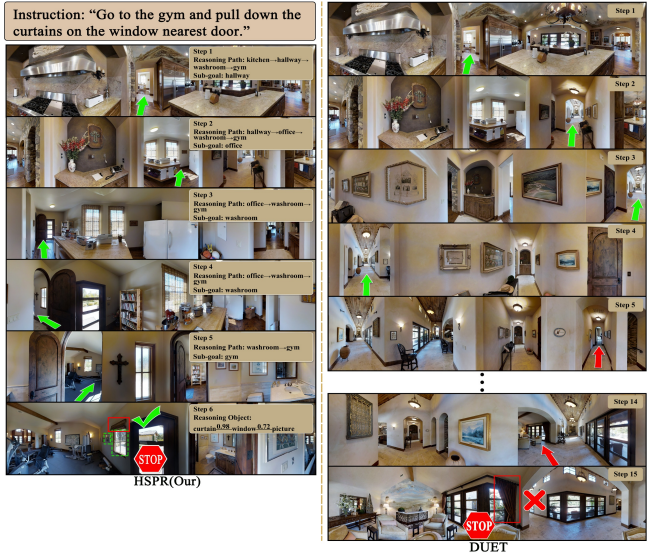

**Fig. 5.** Navigation examples of HSPR and DUET [11]. The left side shows our HSPR method, while the right side represents the baseline DUET.

**Multi-step Reasoning Navigation.** The number of steps in multi-step reasoning refers to the maximum number of node types that the agent will pass through. The impact of reasoning steps on navigation performance is shown in Table VI. As the number of steps increases, the SPL gradually improves, with three-step reasoning achieving the highest SPL value of 38.05%. However, beyond this point, as the number of reasoning steps increases, performance begins to decline. We believe this is because the conditions for the existence of nodes inferred at each step must be satisfied by existence of the nodes inferred in the previous step. Consequently, the nodes inferred in the fourth and fifth steps may be less reliable or even non-existent.

**Score Fusion Policy.** In Table VII, we compare three fusion methods: average fusion, dynamic fusion, and the proposed residual fusion. Compared with the first two methods, the residual fusion improves SR by at least 0.87% and SPL by 0.84%.

*F. Qualitative Results*

Fig. 4 presents the visualized results of the node proximity knowledge obtained through the auxiliary task. It can be observed that the spatial proximity probabilities between many nodes are zero, while some are close to one. This is generally consistent with common sense. Our proposed method ensures the accuracy of these probabilities. In contrast, the spatial proximity probabilities extracted from ConceptNet are mainly distributed within a relatively narrow range (0.4 to 0.55), which provides very limited clues for reasoning optimal paths.

Fig. 5 visualizes the reasoning navigation process of HSPR in an unseen scene within REVERIE [2], comparing it with the DUET. In step 1, the HSPR agent identifies the types of the navigable nodes and plans the optimal path. In step 2, the agent discovers a new type of navigable node (an office) and replans the optimal path. From steps 3 to 6, the agent continues planning and searching for sub-goals, eventually reaching the target. The agent then utilizes spatial proximity knowledge of objects to accurately locate the target object (the curtain). In contrast, the DUET model deviated from the shortest path in step 5, continuing to explore the environment in search of the target, and is ultimately forced to terminate after reaching the maximum navigation steps. This demonstrates that guidance from additional knowledge during navigation is superior to unguided exploration.

## V. CONCLUSION

This paper proposes a Hierarchical Spatial Proximity Reasoning (HSPR) method for the vision-and-language navigation. We design a scene understanding auxiliary task to identify the types of nodes in scenes, explore adjacency relationships, and construct a hierarchical spatial proximity knowledge base during the interaction between the agent and the navigation environment. Based on this, we propose a multi-step reasoning navigation algorithm that infers optimal paths to the target node using proximity knowledge. Our approach achieves the best performance on the REVERIE, SOON, R2R, and R4R datasets compared with the baselines, demonstrating its effectiveness. We plan to apply HSPR to continuous environments in future work.

## APPENDIX

Algorithm 1 provides an overview of the workflow of the HSPR method.

---

**Algorithm 1:** Hierarchical Spatial Proximity Reasoning

**Input:** Instruction $I$, Panoramic views $\mathcal{P}$.
**Output:** Navigation action $\alpha$.

**Pre-training stage:**
1: Construct a hierarchical spatial proximity knowledge base.
2: Perform pre-training tasks.

**Navigation stage:**
  **While** not reached target **do**
1: Update the semantic topological map $\mathcal{G}$.
2: Utilize the visual features $v_i$ of navigable nodes to predict their types, and employ the instruction embedding $I'$ to predict target node type.
3: Use the multi-step reasoning algorithm to obtain the $K$ paths with the highest confidence from the current node to the target node based on node proximity knowledge.
4: Select the optimal path and treat the next intermediate node as a sub-goal.
5: Calculate the global and local node proximity scores $\eta_i^c$ and $\eta_i^f$ for the navigable nodes in $\mathcal{G}$ leading to the sub-goal, and add these to the global and local visual scores $\varepsilon_i^c$ and $\varepsilon_i^f$ predicted by DUET, resulting in the global and local action

---

scores $l_i^c$ and $l_i^f$.

6: Convert the local action score $l_i^f$ into the global action score $l_i'^f$ through residual fusion, and then compute the final action score $l_i$ by summing them with weights, ultimately selecting the action $a$ with the highest score.

**end**

## REFERENCES

[1] P. Anderson et al., "Vision-and-language navigation: Interpreting visually-grounded navigation instructions in real environments," in *Proc. IEEE Conf. Comput. Vis. Pattern Recognit.*, 2018, pp, 3674–3683.

[2] Y. Qi et al., "Reverie: Remote embodied visual referring expression in real indoor environments," in *Proc. IEEE Conf. Comput. Vis. Pattern Recognit.*, 2020, pp, 9982–9991.

[3] X. Gao, Q. Gao, R. Gong, K. Lin, G. Thattai, and G. S. Sukhatme, "Dialfred: Dialogue-enabled agents for embodied instruction following," *IEEE Robot. Automat. Lett.*, vol. 7, no. 4, pp, 10049–10056, Oct. 2022.

[4] P. Anderson et al., "Bottom-up and top-down attention for image captioning and visual question answering," in *Proc. IEEE Conf. Comput. Vis. Pattern Recognit.*, 2018, pp, 6077–6086.

[5] F. Zhu, Y. Zhu, X. Chang, and X. Liang, "Vision-language navigation with self-supervised auxiliary reasoning tasks," in *Proc. IEEE Conf. Comput. Vis. Pattern Recognit.*, 2020, pp, 10012–10022.

[6] D. Fried et al., "Speaker-follower models for vision-and-language navigation," *Adv. Neural Inf. Process. Syst.*, vol. 31, 2018, pp. 3318–3329.

[7] S. Hochreiter and J. Schmidhuber, "Long short-term memory," *Neural comput.*, vol. 9, no. 8, pp, 1735–1780, Feb. 1997.

[8] A. Pashevich, C. Schmid, and C. Sun, "Episodic transformer for vision-and-language navigation," in *Proc. IEEE Int. Conf. Comput. Vis.*, 2021, pp, 15942–15952.

[9] K. Chen, J. K. Chen, J. Chuang, M. Vázquez, and S. Savarese, "Topological planning with transformers for vision-and-language navigation," in *Proc. IEEE Conf. Comput. Vis. Pattern Recognit.*, 2021, pp, 11276–11286.

[10] S. Chen, P. L. Guhur, C. Schmid, and I. Laptev, "History aware multimodal transformer for vision-and-language navigation," *Adv. Neural Inf. Process. Syst.*, vol. 34, 2021, pp, 5834–5847.

[11] S. Chen, P. L. Guhur, M. Tapaswi, C. Schmid, and I. Laptev, "Think global, act local: Dual-scale graph transformer for vision-and-language navigation," in *Proc. IEEE Conf. Comput. Vis. Pattern Recognit.*, 2022, pp, 16537–16547.

[12] A. Vaswani, N. Shazeer, N. Parmar, J. Uszkoreit, L. Jones, A. N. Gomez, Ł. Kaiser, and I. Polosukhin, "Attention is all you need," *Adv. Neural Inf. Process. Syst.*, vol. 30, 2017, pp. 5998–6008.

[13] C. Gao, J. Chen, S. Liu, L. Wang, Q. Zhang, and Q. Wu, "Room-and-object aware knowledge reasoning for remote embodied referring expression," in *Proc. IEEE Conf. Comput. Vis. Pattern Recognit.*, 2021, pp, 3064–3073.

[14] X. Li, Z. Wang, J. Yang, Y. Wang, and S. Jiang, "Kerm: Knowledge enhanced reasoning for vision-and-language navigation," in *Proc. IEEE Conf. Comput. Vis. Pattern Recognit.*, 2023, pp, 2583–2592.

[15] R. Speer, J. Chin, and C. Havasi, "Conceptnet 5,5: An open multilingual graph of general knowledge," in *Proc. Assoc. Adv. Artif. Intell.*, 2017, pp, 4444–4451.

[16] R. Krishna et al., "Visual genome: Connecting language and vision using crowdsourced dense image annotations," *Int. J. Comput. Vis.*, vol. 123, pp, 32–73, Feb. 2017.

[17] V. Jain, G. Magalhaes, A. Ku, A. Vaswani, E. Ie, and J. Baldridge, "Stay on the path: Instruction fidelity in vision-and-language navigation," in *Proc. Assoc. Comput. Linguistics.*, 2019, pp, 1862–1872.

[18] A. Ku, P. Anderson, R. Patel, E. Ie, and J. Baldridge, "Room-across-room: Multilingual vision-and-language navigation with dense spatiotemporal grounding," in *Proc. Conf. Empirical Methods Natural Lang. Process.*, 2020, pp, 4392–4412.

[19] F. Zhu, X. Liang, Y. Zhu, Q. Yu, X. Chang, and X. Liang, "Soon: Scenario oriented object navigation with graph-based exploration," in *Proc. IEEE Conf. Comput. Vis. Pattern Recognit.*, 2021, pp, 12689–12699.

[20] D. S. Chaplot, R. Salakhutdinov, A. Gupta, and S. Gupta, "Neural topological slam for visual navigation," in *Proc. IEEE Conf. Comput. Vis. Pattern Recognit.*, 2020, pp, 12875–12884.

[21] H. Wang, W. Wang, W. Liang, C. Xiong, and J. Shen, "Structured scene memory for vision-language navigation," in *Proc. IEEE Conf. Comput. Vis. Pattern Recognit.*, 2021, pp, 8455–8464.

[22] R. Hu, D. Fried, A. Rohrbach, D. Klein, T. Darrell, and K. Saenko, "Are you looking? grounding to multiple modalities in vision-and-language navigation," 2019, *arXiv*:1906.00347.

[23] Y. Hong, Q. Wu, Y. Qi, C. Rodriguez-Opazo, and S. Gould, "Vln BERT: A recurrent vision-and-language BERT for navigation," in *Proc. IEEE Conf. Comput. Vis. Pattern Recognit.*, 2021, pp, 1643–1653.

[24] X. Wang et al., "Reinforced cross-modal matching and self-supervised imitation learning for vision-language navigation," in *Proc. IEEE Conf. Comput. Vis. Pattern Recognit.*, 2019, pp, 6629–6638.

[25] H. Wang, W. Wang, T. Shu, W. Liang, and J. Shen, "Active visual information gathering for vision-language navigation," in *Proc. Eur. Conf. Comput. Vis.*, 2020, pp, 307–322.

[26] J. Y. Koh, H. Lee, Y. Yang, J. Baldridge, and P. Anderson, "Pathdreamer: A world model for indoor navigation," in *Proc. IEEE Int. Conf. Comput. Vis.*, 2021, pp, 14738–14748.

[27] G. A. Sigurdsson, J. Thomason, G. S. Sukhatme, and R. Piramuthu, "RREx-BoT: Remote Referring Expressions with a Bag of Tricks," in *Proc. IEEE Int. Conf. Intell. Robots Syst.*, 2023, pp, 5203–5210.

[28] H. Tan, L. Yu, and M. Bansal, "Learning to navigate unseen environments: Back translation with environmental dropout," in *Proc. Conf. North Amer. Chapter Assoc. Comput. Linguistics.*, 2019, pp, 2610–2621.

[29] H. Huang, V. Jain, H. Mehta, A. Ku, G. Magalhaes, J. Baldridge, and E. Ie, "Transferable representation learning in vision-and-language navigation," in *Proc. IEEE Int. Conf. Comput. Vis.*, 2019, pp, 7403–7412.

[30] J. Lu, D. Batra, D. Parikh, and S. Lee, "Vilbert: Pre-training task-agnostic visiolinguistic representations for vision-and-language tasks," *Adv. Neural Inf. Process. Syst.*, vol. 32, 2019, pp, 13–23.

[31] P.L. Guhur, M. Tapaswi, S. Chen, I. Laptev, and C. Schmid, "Airbert: In-domain pre-training for vision-and-language navigation," in *Proc. IEEE Int. Conf. Comput. Vis.*, 2021, pp, 1634–1643.

[32] C. W. Kuo, C. Y. Ma, J. Hoffman, and Z. Kira, "Structure-Encoding Auxiliary Tasks for Improved Visual Representation in Vision-and-Language Navigation," in *Proc. IEEE Winter Conf. Appl. Comput. Vis.*, 2023, pp, 1104–1113.

[33] J. Ye, D. Batra, E. Wijmans, and A. Das, "Auxiliary tasks speed up learning point goal navigation," in *Proc. Conf. Robot Learn.*, 2021, pp, 498–516.

[34] C. Y. Ma et al., "Self-monitoring navigation agent via auxiliary progress estimation," in *Proc. Int. Conf. Learn. Representations*, 2019.

[35] A. Majumdar, A. Shrivastava, S. Lee, P. Anderson, D. Parikh, and D. Batra, "Improving vision-and-language navigation with image-text pairs from the web," in *Proc. Eur. Conf. Comput. Vis.*, 2020, pp, 259–274.

[36] H. Tan and M. Bansal, "Lxmert: Learning cross-modality encoder representations from transformers," in *Proc. Conf. Empirical Methods Natural Lang. Process.*, 2019, pp. 5100–5111.

[37] J. Devlin, M. W. Chang, K. Lee, and K. Toutanova, "Bert: Pre-training of deep bidirectional transformers for language understanding," in *Proc. Conf. North Amer. Chapter Assoc. Comput. Linguistics.*, 2019, pp, 4171–4186.

[38] X. Lin, G. Li, and Y. Yu, "Scene-intuitive agent for remote embodied visual grounding," in *Proc. IEEE Conf. Comput. Vis. Pattern Recognit.*, 2021, pp, 7036–7045.

[39] A. Dosovitskiy et al., "An image is worth 16×16 words: Transformers for image recognition at scale," in *Proc. Int. Conf. Learn. Representations*, 2020.

[40] Y. Qi, Z. Pan, S. Zhang, A. Van Den Hengel, and Q. Wu, "Object-and-action aware model for visual language navigation," in *Proc. Eur. Conf. Comput. Vis.*, 2020, pp, 303–317.